\documentclass{article}

\usepackage{microtype}
\usepackage{graphicx}
\usepackage{subcaption}
\usepackage{booktabs} 
\usepackage{listings}

\usepackage[utf8]{inputenc}
\usepackage[english]{babel}

\usepackage[T1]{fontenc}
\usepackage[T1]{fontenc}
\usepackage{hyperref}
\usepackage[newfloat,cachedir=minted,frozencache]{minted}
\newcommand{\ic}[1]{\mintinline{Python}{#1}}

\usepackage[accepted]{mlsys2020}

\mlsystitlerunning{Using Python for Model Inference}
\usepackage{inconsolata}

\begin{document}

\twocolumn[
\mlsystitle{Using Python for Model Inference in Deep Learning}



\mlsyssetsymbol{equal}{*}

\begin{mlsysauthorlist}
\mlsysauthor{Zachary DeVito}{fair}
\mlsysauthor{Jason Ansel}{fair}
\mlsysauthor{Will Constable}{fair}
\mlsysauthor{Michael Suo}{fair}
\mlsysauthor{Ailing Zhang}{fair}
\mlsysauthor{Kim Hazelwood}{fair}
\end{mlsysauthorlist}

\mlsysaffiliation{fair}{Facebook AI Research, Menlo Park, California, USA}
\mlsyscorrespondingauthor{Zachary DeVito}{zdevito@cs.stanford.edu}

\mlsyskeywords{Machine Learning, MLSys}

\vskip 0.3in

\begin{abstract}
Python has become the de-facto language for training deep neural networks, coupling a large suite of scientific computing libraries with efficient libraries for tensor 
computation such as PyTorch~\cite{pytorch} or TensorFlow~\cite{tensorflow}. However, when models are used for inference they are typically extracted from Python as TensorFlow graphs or TorchScript~\cite{torchscript} programs in order to meet performance and packaging constraints. The extraction process can be time consuming, impeding fast prototyping. We show how it is possible to meet these performance and packaging constraints while performing inference in Python. In particular, we present a way of using multiple Python interpreters within a single process to achieve scalable inference and describe a new container format for models that contains both native Python code and data. This approach simplifies the model deployment story by eliminating the model extraction step, and makes it easier to integrate existing performance-enhancing Python libraries. We evaluate our design on a suite of popular PyTorch models on Github, showing how they can be packaged in our inference format, and comparing their performance to TorchScript. For larger models, our packaged Python models perform the same as TorchScript, and for smaller models where there is some Python overhead, our multi-interpreter approach ensures inference is still scalable.

\end{abstract}
]



\printAffiliationsAndNotice{}  

\section{Introduction}
\label{introduction}
Over the past few years, Python has become the de-facto language for training deep learning models with major deep learning frameworks such as PyTorch~\cite{pytorch}, TensorFlow~\cite{tensorflow}, MxNet/TVM~\cite{mxnet, tvm} and JAX~\cite{jax} primarily providing Python interfaces. Python's ease of writing code and distributing it via package managers such as conda~\cite{conda} make it easy to share libraries and research results. Its ecosystem of libraries for data science such as numpy~\cite{numpy},  pandas~\cite{pandas}, and Jupyter notebooks~cite{jypter} make analyzing training results easy.

However, once a model is trained, using the model resembles traditional software engineering more than data science. It is desireable to have a stand-alone and reproducible artifact that can run the model independently from its training harness. These artifacts are typically then used as part of a service such as TorchServe~\cite{torch_serve} that manages inference of the model, or as a library call made within an application.

The process of extracting a model into a stand-alone artifact can be time consuming. Approaches such as partial evaluating the model~\cite{autograph}, or compiling a subset of Python to a stand alone language (TorchScript~\cite{torchscript}) often require manual intervention to refactor the model into a more limited programming model than the one offered by Python and are a common source of confusion for users of deep learning frameworks. Often features that made the model easy to work with in Python, such as dynamic typing or object-oriented configuration of model layers have to be removed in this step. Frequently the user extracting the model is not the same as the original model author, making the kind of whole-program refactorings needed to get the program extracted difficult.

We offer an alternative workflow by revisiting the assumption that it is not feasible to run Python for model inference. Reticence to running Python as a production language stems from problems of running Python as a platform for web services such as Django~\cite{django}. In these settings ,there are lots of small objects and millions of lines of code. Deep learning models have vastly smaller amounts of code and fewer but bigger objects. In many cases it is possible to simply use the existing CPython interpreter as a platform for model inference.  We show how it is possible to organize CPython such that multiple independent interpreters can co-exist in the same process. We also develop a new hermetic packaging format that makes it possible to easily create self-contained model artifacts from existing Python code and trained weights. In particular, we present the following contributions:

\begin{itemize}
\item An analysis of the challenges of using Python for model inference.
\item A scalable approach to serving models using multiple Python interpreters in the same process that can share weight tensors.
\item A hermetic packaging format for Python, \ic{torch.package}, that can create self-contained archives of model code and weights.
\item A library-based C++ API, \ic{torch::deploy}, for using our multi-interpreter approach.
\item An evaluation of the performance of this approach compared to TorchScript, showing model performance for many models is nearly the same as TorchScript, and for smaller models with Python overhead, our multi-interpreter approach is still scalable.
\end{itemize}

\section{Background}
In most deep learning frameworks, model inference is accomplished by first exporting the model into a stand-alone format. TensorFlow represents models using files containing protocol buffers describing the graph of the model. In TensorFlow 2.0, an eager mode was introduced which runs directly in Python. An export process, autograph~\cite{autograph} can partially evaluate the eager-mode program to derive a graph that can be exported. PyTorch's default mode is eager evaluation in Python. To export a model a model users can trace this execution (if it does not contain control-flow), producing an exportable ONNX graph that can be loaded from other inference frameworks~\cite{onnx}. For more complicated models, users can convert the model to TorchScript~\cite{torchscript}, a process that requires writing the model in a statically-typed subset of the Python language. A separate TorchScript interpreter can be used to deploy the model. In any of these approaches, the developer may need to spend significant time refactoring a Python-based model to be ready for export. For instance, the user might need to refactor the model to typecheck TorchScript's static type system, or to remove aggregate data types that cannot be represented in a TensorFlow graph.

While it is possible to simply run inference in Python, deep learning frameworks have not previously recommended it. One possible explanation is the history of using Python in production settings. Python is a popular language for writing webservers with Django~\cite{django} being the most commonly used libraries. However, as individual web servers grow more complex, they face problems with the performance of the standard CPython implementation, leading to efforts to improve Python's performance. This includes Shedskin~\cite{shedskin} a project developed at Google, Pyston~\cite{pyston} developed at Dropbox, and modifications to the CPython interpreter used by Instagram~\cite{instagram}.  Eventually, developers sought other approaches to their problem such as migrating to other languages such as Go~\cite{pyston}.  While faster Python interpreters such as PyPy~\cite{pypy} exist, they are not commonly used because they do not have the same extension framework as CPython, making the use of common libraries like numpy or TensorFlow or PyTorch difficult. 

\section{Challenges using CPython}
Running CPython for deep learning inference is met with skepticism due to these well known challenges in efficiently running Python code using the CPython interpreter. Naively running PyTorch Python programs for inference would run into similar scaling problems, so this initial skepticism is warranted. However by breaking down the challenges in running Python code efficiently, we can separate the problems we have to address for deep learning inference from ones that can be mitigated in other ways.
\subsection{For performance}
\begin{figure}
\includegraphics[width=\linewidth]{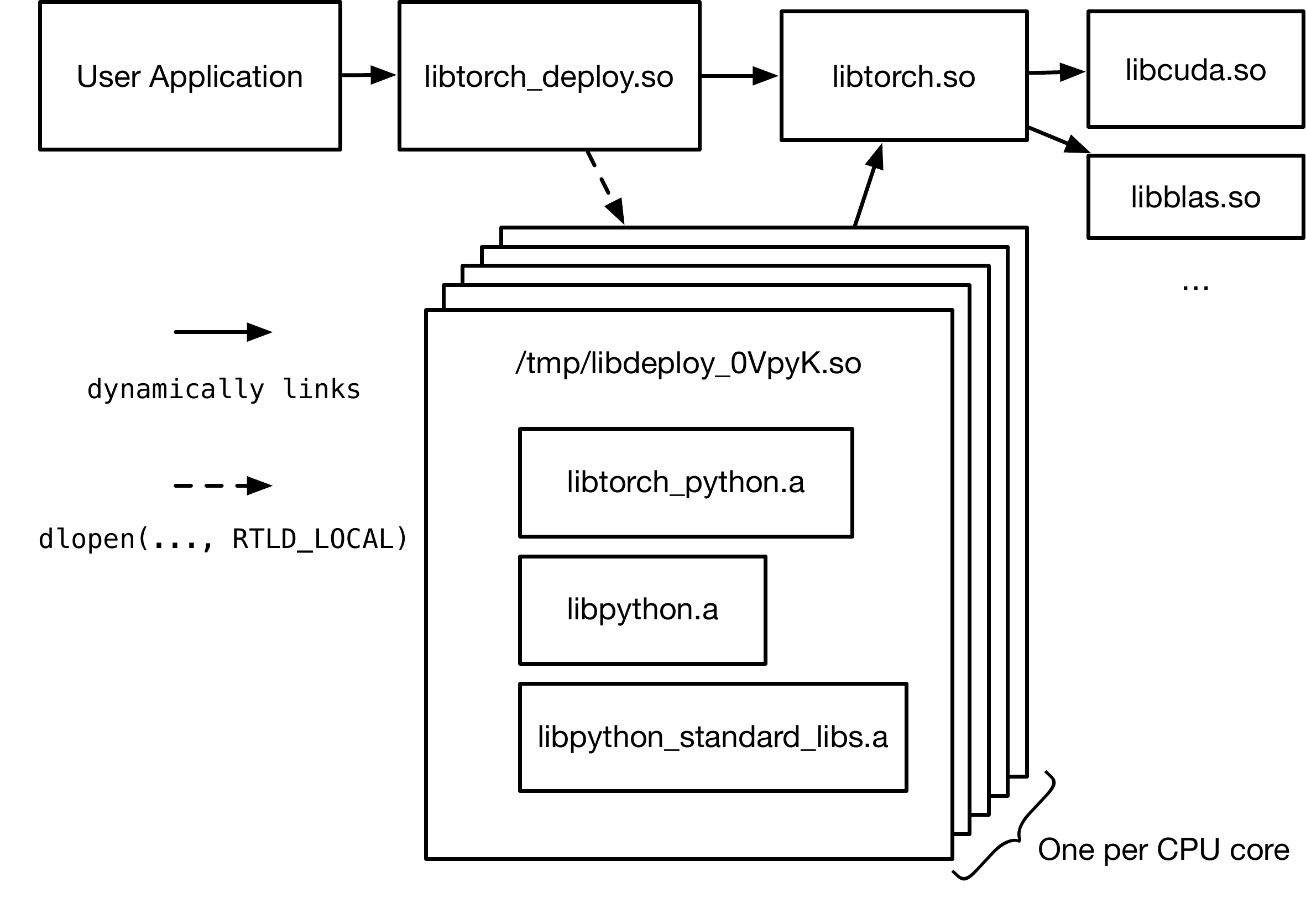}
\caption{A visualization of the library dependencies of our approach. Our library, \ic{libtorch_deploy.so} presents a public API to the user's application. Internally, it supports loading multiple versions of the Python interpreter by encapsulating them in a shared library object which is copied to a temporary location (\ic{/tmp/libdeploy...}) before being loaded privately using \ic{dlopen} with \ic{RTLD_LOCAL}. This library contains all direct uses of the Python C API from PyTorch (\ic{libtorch_python.a}). The process of copying and loading the library can be repeated to create multiple interpreters. These copies all dynamically link against the same copy of Torch's C++ implementation \ic{libtorch.so}, so they can share this code.}
\label{libraries}
\end{figure}
\paragraph{Global interpreter lock} The most widely understood challenge with using CPython is that its design requires a single global context, and only a single instance of a Python interpreter can use that context at once. This constraint is enforced using a a global interpreter lock (GIL) that must be held by the thread using the Python context. This design simplifies the implementation of core infrastructure such as garbage collection and the memory format of objects. Languages that do allow for multiple threads to run at once have to carefully reason about the implementation, and provide a detailed contract (e.g. the Java memory model~\cite{java_memory_model}) to users of threads about how objects behave when accessed in multiple threads. Attempts to remove the GIL~\cite{attempts_to_remove_GIL} have faced challenges in providing an implementation that does not degrade single-threaded performance, defining a reasonable memory model, and then updating library code to work within this memory model. 

For server applications handling multiple requests at once such as a server for deep learning inference, the Python interpreter becomes a bottleneck to scaling and experiments in our evaluation section confirm this is a significant issue for some of our models. 
Recent proposals for Python will eventually allow multiple separate interpreters~\cite{python_interps}. While objects cannot be shared among interpreters, each interpreter would have its own GIL, allowing for parallel execution. However, for practical purposes it will be a some time before this becomes a viable option because popular extensions in the Python ecosystem such as numpy, scipy, or PyTorch are coded with the assumption that there is a single Python context and will need to be modified before they can be used with multiple interpreters.

A popular workaround is to use the Python \ic{multiprocessing} library which provides primitives to allow multiple Python \emph{processes} to cooperate. In this model each CPU core might have its own Python process. Since each process has its own Python state, the processes cannot directly share Python objects. Instead, inter-process communication is implemented using Python's \ic{pickle} protocol to serialize and deserialize objects that are passed between interpreters. Non-python data, such as the numbers inside a PyTorch Tensor can be shared using the OS's facilities to share memory between processes. Several aspects of this setup are not ideal. First, having to manage a process per CPU core (frequently up to 64 cores for big machines) can be cumbersome. Tools like debuggers often do not understand that the collection of processes is a single application, making it hard to use these tools. Task management and logging become more challenging as well, especially when these interpreters are only a small part of a bigger application. Second, sharing non-python resources is not always possible without special OS support. For instance, memory needs to be specially allocated to share it across processes. In PyTorch this manifests as needing to copy tensors to make them sharable. Sharing special memory such as CUDA allocated memory or memory allocated by libraries for other custom accelerator chips is also difficult or impossible.

\paragraph{A highly customizable object model} 
CPython objects can be defined using a C extension API. This API exposes the details of the object layout directly to extension writers and allows nearly every aspect of objects to be defined in an object-specific way including custom behavior for common operators and how they are scanned for garbage collection. Common types are also defined using this API. This makes it challenging to use JIT compilation techniques to accelerate Python that have been successful in other dynamically-typed scripting languages such as Javascript~\cite{jit_in_javascript}. Attempts to compile the interpreter bytecodes in Python yield little speedup because most of the time is spent in functions like \ic{PyObject_Add} that have to look up how to dynamically dispatch an add using details of the object model~\cite{python_interpreter_overhead1, python_interpreter_overhead2}.  Since most of the complexity is in the object model, attempts like PyPy~\cite{pypy} to restructure the object model allow JITs to accelerate the speed of Python code, but since they do not expose the same extension API, the common parts of the Python ecosystem such as numpy or PyTorch do not natively work. The result is that the speed of Python on small micro-benchmark programs such as those found in the programming language benchmark game~\cite{benchmark_game} are a median of 37 times slower than native execution in C and 13 times slower than Javascript.

\paragraph{Memory Management}
CPython reference counts an object each time an object is saved in the heap or interpreter stack. This design requires a lot more writes to occur than a pure garbage collector, which can decrease the performance of caches, and it complicates efforts to remove the interpreter lock since reference counting would then need to become more costly atomic operators. Since reference counting would leak cycles, Python also includes a cycle collector to find unreachable cycles. While it runs less frequently than a fully-deferred garbage collector, latency can still become variable for programs that create a lot of cycles for small objects. In practice for server uses of Python, this has lead large deployments like Instagram to use tricks like forking the Python process from a clean start state and disabling the garbage collector entirely~\cite{instagram}.

\subsection{For deep learning inference}
Deep learning inference has its own unique properties that present additional challenges to using CPython.
\paragraph{Model Packaging} Model code, unlike normal source code, is closely coupled with large parameters (model weights) normally stored as separate binary data during the training process. For the model to work correctly, the code that executes the model has to be the same as when it was trained. Small changes might cause losses in accuracy because of the way gradient descent tunes the weights for any particular quirks of the implementation. Existing packaging solutions typically store the model code and weights together. For instance, TensorFlow stores serializes the graph and its weights into a protobuf format~\cite{tensorflow}. Similarly TorchScript~\cite{torchscript} uses a zip file format that contains compiled TorchScript code (extracted from the original model code), and embedded weight files. 

Python code is typically stored separate from data, either as source files in some source control system, or in versioned packages in managers such as conda~\cite{conda}. Neither approach is ideal for packaging models. Using source control systems to store large binaries, sometimes including gigabytes of weights, is not practical because they retain all versions of the data.  Furthermore, models are frequently produced using a search over some hyperparameter space with training that may include manual tweaking of the model code. Asking modellers to upstream those tweaks into a shared code repository is not something that is currently expected, even if it might be good software engineering practice in the future. 

Storing Python models as pip or conda packages allows the storage of additional binary data like weights, but also requires that models provide globally unique names for themselves and any components that they include. This can become cumbersome since many models share very similarly named things (e.g. a ResNet block) that nevertheless may have slight differences. This would make it difficult for a multi-tenant inference server to manage a suite of similar but possibly mutually incompatible Python packages.

\paragraph{Embedability}
Model inference is commonly done in two ways: as a function call within a user's existing application (such as a desktop app, or language translation service), or as a remote procedure call to a dedicated server than runs many different kinds of models such as TorchServe~\cite{torch_serve}. In both cases, the application does not care about the details of the implementation of the model, only that it can be run efficiently and return a result.  

A single global Python instance like that provided by CPython makes it difficult to embed libraries that use the Python interpreter in other applications. For instance, assume we provide an inference library \ic{libtorchserving.so} that internally uses the Python interpreter. If the application itself also uses Python in some way then the Python environment in our library will leak into the environment of the application, potentially leading to name conflicts such as having two different versions of the same Python library among other unexpected behaviors. Problems with Python multiprocessing are exasperated in this setup as well since the application may already have its own pool of worker threads. That pool might then end up contending with the multiprocessing pool in ways that degrade performance. 

Python itself also typically requires its standard library to be present on the filesystem of the machine in order for most commonly used Python modules to function correctly. For an embedded inference library needing to also ship a set of discover-able Python files presents a challenge for integration into arbitrary applications.

\section{Approach}
\begin{figure}[h!]
\begin{minted}[fontsize=\scriptsize]{cpp}
using namespace torch::deploy;
void main() {
  // an interpreter manager load balances requests across
  // Python interpreters.
  InterpreterManager manager(/*n_interpreters=*/8);
  
  // Open a package saved with torch.package
  Package package = manager.load_package("mobile_net_v3");
  
  // get the pickled model object out of the package, and turn it 
  // into a MovableObject which can be loaded onto any interpreter.
  MovableObject model = package.load_pickle("model", 
                                            "model.pkl");
  
  #omp parallel for
  for(size_t i = 0; i < 8; ++i) {
    torch::Tensor = input_image = torch::rand(1, 3, 224, 224);
    // run the model from multiple threads,
    // the manager will load balance the requests across
    // interpreters, loading the MovableObject onto each interpreter
    // the first time it is required
    torch::Tensor result = model({input_image}).toTensor();
  }
  
  // A lower-level API can be used to directly interact with Python.
  // Here we use it to wrap the model in a GPUInferenceModel which
  // handles moving inputs to/from the GPU.
  MovableObject gpu_model;
  {
    // start an interaction with one Python interpreter to 
    // do model loading
    InterpreterSession s = package.acquire_session();
    // s.self is the torch.package.PackageImporter object
    // which we can directly interact with using Python bindings:
    PyObj cpu_model = s.self.attr("load_pickle")({"model", 
                                                  "model.pkl"});
    PyObj GPUModel = s.global("gpu_inference", "GPUModel");
    PyObj gpu_model_py = GPUModel({cpu_model});
    // turn the loaded Python model into an object that can be 
    // moved across different python interpreters, using pickle:
    gpu_model = s.create_movable(gpu_model_py);
  }
  
  
  #omp parallel for
  for(size_t i = 0; i < 8; ++i) {
    // during inference we can also acquire a handle to the Python
    // interpreter to do more than just call forward:
    InterpreterSession s = gpu_model.acquire_session();
    torch::Tensor = input_image = torch::rand(1, 3, 224, 224);
    PyObj output_image = s.self({input_image});
    PyObj post = s.global("post_process", "PostProcessImage");
    PyObj post_processed = post({output_image});
    torch::Tensor result = post_processed.toIValue().toTensor()  
  }
}
    \end{minted}
    \caption{An example showing how to use the \ic{torch::deploy} C++ API to run Python inference.}
    \label{api}
\end{figure}

To use Python models for inference, we need to address the performance and packaging challenges presented in the previous section. 

\subsection{Mitigating performance factors}

Some aspects of performance are mitigated by the unique properties of deep learning models compared to general purpose Python programs. 

A Python webserver such as Django might have many small objects representing the components of a web application. Each object might have 10s of fields, stored as Python objects in a dictionary. Working with these objects is slow because of the Python overheads of unpacking each dynamically typed object. In contrast, PyTorch programs primarily work with Tensors that typically contain greater than 256 elements and frequently contain many more. 

For instance, the size the activation Tensors in an unbatched ResNet50 ranges from 25k to 100k floats. This means that there are fewer Python objects overall compared to general Python programs and the objects themselves are much larger.  

The difference in object number and size has a number of mitigating effects on Python performance. First, programs spend relatively little time in the Python interpreter and its object model. As an example, if we run an unbatched ResNet50 model on the CPU, replacing all tensors with dummy objects that do no compute, we find that Python execution only takes 13\% of runtime. 

Having fewer objects also makes memory management less concerning. Referencing counting, while expensive for small objects, occurs infrequently in deep learning programs due to the larger object size. 


Having few but large tensor objects also opens up new possibilities for data sharing across inference. For instance, it is relatively inexpensive to copy an entire PyTorch model object as long as the Tensors are still shared across copies.

Another mitigating property of model inference is the inherent parallelism of handling multiple inference requests in parallel. Requests can be fulfilled on entirely in independent threads, possibly working on different models. Within a model, data-parallelism can be exploited by doing batched inference. General uses of Python microprocessing face the overhead of pickling and unpickling Python objects to pass them between processes. In model inference, no data needs to be exchanged between requests so the speed of transferring Python objects is less of a concern.

\subsection{Strategy}

The unique properties of model inference means that we can work around Pythons relative slow object and memory management model. However, we still need address the global interpreter lock, the embedability of the interpreter itself, and the challenges of packaging model parameters with Python code. To address the GIL and embedability, we propose a way of building the CPython library so that it can be loaded privately multiple times in the same process. To address packaging, we present a new container format for Python code and parameter data along with a way of loading these containers that ensures its code does not conflict with other containers.

\subsection{An embedded CPython}
\label{multi_python}
Our approach for working around the GIL is to create a version of the CPython interpreter that can be loaded privately multiple times in a single process. On top of this interpreter, we build a model serving infrastructure that can hand off requests to different interpreters in order to scale to multiple threads without GIL contention. We first look at how we construct the private interpreter and then how we use it to build an inference engine.

To construct an interpreter that can be loaded independently, we create a special shared library \ic{libinterp.so} that contains the CPython implementation \ic{libpython.a}, any extensions modules that bind to it such as PyTorch's Python bindings(\ic{libtorch_python.a}), and a small API for transferring data into and out of the private interpreter. These components are linked together statically in the shared library and a linker script is used to mark the visibility of the symbols in the library as hidden. Dependencies that do not use the Python APIs such as PyTorch's C++ library (\ic{libtorch.so}) are linked against dynamically. An inference application can then load a copy of interpreter using \ic{dlopen}. Passing the \ic{RTLD_LOCAL} flag ensures that the symbols loaded in this library will not be visible to the rest of the application. 

However, this packaging only provides a single copy of Python because the normal shared library system will ensure any particular shared library is only loaded once. To work around this, we first \emph{copy} the shared library to a temporary file on the file system before loading with \ic{dlopen}, by doing this we ensure we get unique copies of the library. On load, the library will resolve its shared library dependencies to the symbols already loaded in the process ensuring. Figure~\ref{libraries} illustrates the result. Everything loaded in \ic{libdeploy.so} will have multiple copies, but shared library dependencies will be resolved globally ensuring there is only a single copy of libraries like \ic{libtorch.so}.

To ensure that the embedded Python library has access to the Python standard library, we pre-compile the library to Python bytecodes and then embed those bytecodes into the data section of our library, ensuring that the interpreters will not need access to the filesystem.

Because the \ic{libdeploy.so} privatizes the Python APIs, the application cannot directly access the Python state of the interpreters. Instead we provide a minimal C++ API, \ic{torch::deploy} for getting data into and out of the private interpreter. Example uses of the API are shown in Listing~\ref{api}.

We represent each copy of the interpreter as a \textbf{Interpreter} object in C++. An \textbf{InterpreterManager} is a pool of interpreters that can be handed out to different threads when requested. For instance, for an \ic{N} core machine, we create pool of \ic{N} interpreters, treating each interpreter as a resource that can be acquired when the inference library needs to run Python code to either load or run a model. Having one interpreter per real hardware thread ensures we do not end up with a situation where we have a CPU ready to do work but no free interpreter to perform the job.

A \textbf{Package} object is a C++ wrapper around the \ic{torch.package} format described in the next section and is used to load packaged models onto the private interpreters.
To load a model, the user of the library will typically open a package file and load a model from it. This process uses a Python interpreter which is internally acquired from the InterpreterManager. While it would be possible to use the package API to load the model multiple times onto each interpreter, it is not optimal for two reasons. First, loading the model multiple times would result in different copies of the weight tensors. For large models, it is not possible to fit many copies of the model in GPU memory. Second, it is frequently the case that users will want to do some pre-processing to each model, such as wrapping it in a container to manage the movement of data to the GPU, or to connect the model with data post-processing code. Re-doing this preprocessing would be expensive.

To allow models to share weights and pre-processing, we load the model on a single interpreter and then use Python's \ic{pickle} protocol to move the model's Python objects to other interpreters as needed. Because \ic{libtorch.so} is global to the process, we can share the Tensor data between interpreters by customizing the pickling process. The \ic{multiprocessing} framework in Python uses a similar approach but unlike \ic{multiprocessing} the shared data lives in a single process so we do not have to make any special OS calls to create this shared relationship and any use of accelerator libraries such as CUDA works without further modifications.

We abstract this process of loaded and then moving a model in our C++ API as a \textbf{MovableObject} that can be created from a Python object after it is loaded. When the application makes an inference call, an interpreter is allocated out of the pool and we check to see if it has loaded the MovableObject into its Python state. If not, it is loaded from the pickled copy. The loaded object is then used to run inference.

To make an interference call, or otherwise directly interact with an instance of the Python interpreter, we provide a \textbf{InterpreterSession} object. This object acts as a resource guard in C++ that permits direct access to the Python interpreter using \textbf{PyObj} handles which are valid only for the lifetime of the session. A minimal API is exposed to access global values, call functions, and translate primitive values (including Tensors) between C++ and Python. The second example in Figure~\ref{api} shows an example of using this session API directly but typically users will use syntax sugar encapsulates these details.

The inference library intentionally does not have a thread pool. Instead, we expect the inference application to call the library from multiple request threads if desired. The InterpreterManager, rather than being a pool of worker threads, serves as a load balancer for handing out interpreters. This choice is subtle but important. It is typical for applications to be doing other work such as serving models in other formats such as TorchScript or handling non-inference work. It is likely the application already has a threadpool which would fight with an internal threadpool in the inference library.

We also intentionally load multiple models onto a single interpreter rather than use one interpreter per model. This is because each interpreter is relatively expensive. It requires a copy the python interpreter and PyTorch's bindings be made in RAM, and it requires the initialization of Python and PyTorch. By limiting the total number of interpreters to the number of CPU threads, we ensure that we have enough available parallelism to avoid GIL contention but bound the amount of resources the interpreters consume. Instead our packaging format ensures that multiple models do not interfere with each other.

\subsection{Hermetically packaging models}
We propose a new way to package code and model data together to create a self-contained archive for models. Conceptually, it extends Python's existing pickling format to also store the code depended on by the pickled objects. We couple this with a \emph{hermetic} importer that loads archives without polluting the global module table, and ensures that the loaded code only depends on explicitly-declared external dependencies.

\begin{figure*}[t]
     \begin{subfigure}[b]{0.3\textwidth}
        \includegraphics[width=\textwidth]{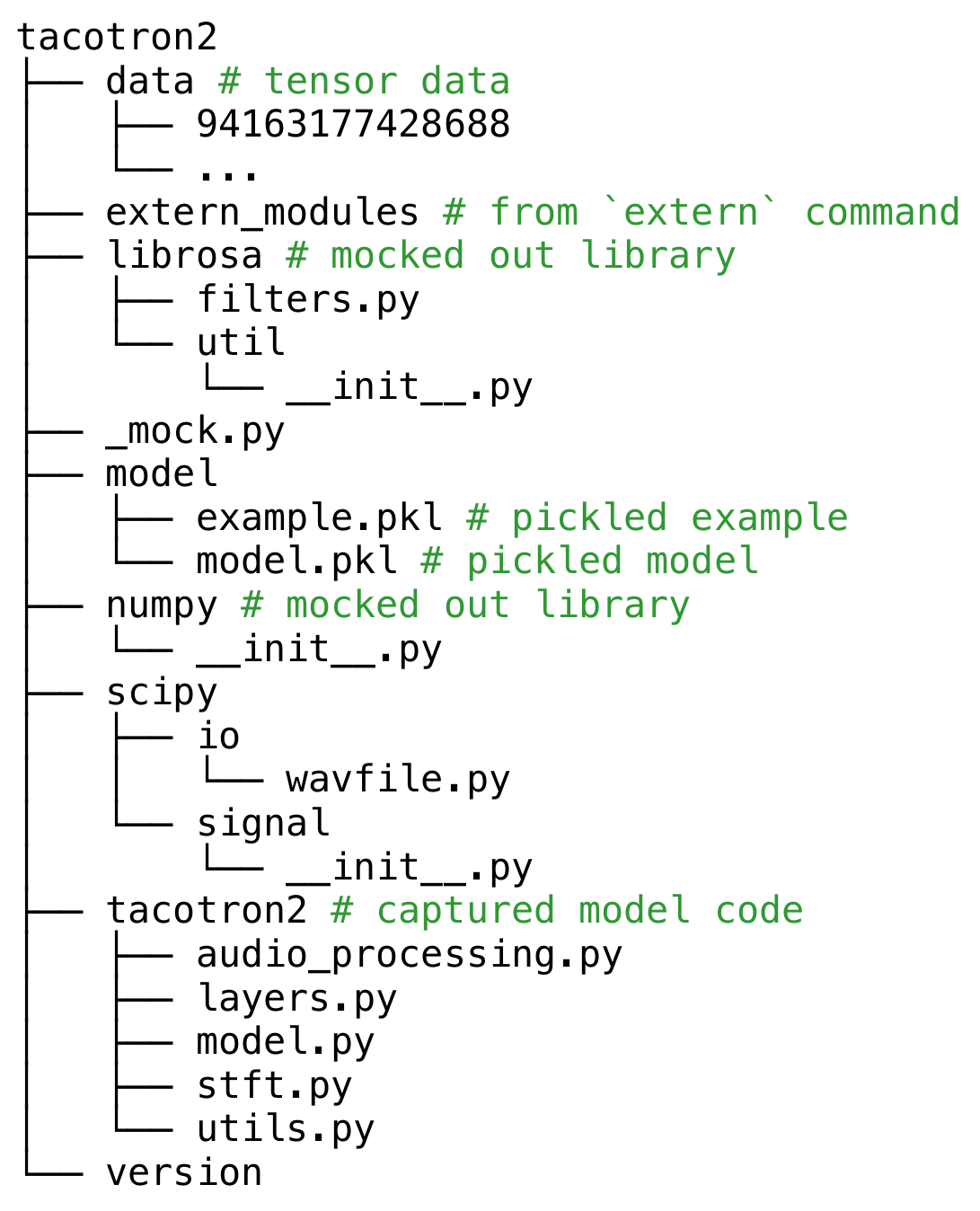}
     \hfill
     
         \caption{Packaged model structure}
         \label{packaging_zip}
     \end{subfigure}
     \begin{subfigure}[b]{0.35\textwidth}
\begin{minted}[fontsize=\scriptsize]{python}
from torch.package import PackageExporter

model, example_input = load_tacotron2()

with PackageExporter('tacotron2') as e:
  # Configure how to export the source code
  e.extern(['torch.**'])
  # instead of saving the source code for the 
  # torch libraries, let the package link to
  # the libraries of the loading process.

  # Replace these modules with mock 
  # implementations. They are not 
  # actually used.
  e.mock(['numpy', 'librosa.**', 'scipy.**'])


  e.save_pickle('model', 'model.pkl', model)
  # dependency resolution will walk 
  # the pickled objects and find all the 
  # required source files

  # also save example tensor for testing
  e.save_pickle('model', 'eg.pkl', 
                example_input)
\end{minted}
         \caption{Model export}
         \label{example_export}
     \end{subfigure}
     \begin{subfigure}[b]{0.35\textwidth}
\begin{minted}[fontsize=\scriptsize]{python}
from torch.package import PackageImporter

i = PackageImporter('tacotron2')

# code for the model is loaded from the model_file
# rather than the normal import system, except
# where packages are marked as extern.
model = i.load_pickle('model', 'model.pkl')
example_inputs = i.load_pickle('model', 'eg.pkl')
# test the model
model(*example_inputs)
\end{minted}
         \caption{Model import}
         \label{importing_api}
     \end{subfigure}
        \caption{An example of the structure of our model packaging format and the code used to export it and import it.}
        \label{fig:three graphs}
\end{figure*}

\subsubsection{Format}
The on-disk format is a zip-file archive. PyTorch already stores its serialized data in this form in order to keep the tensor data as separate mmap-able files. Similarly, Python supports loading code from zip archives such as the "egg" format~\cite{python_egg}. The archive stores several kinds of files:
\begin{itemize}
    \item Python source files, laid out in a package hierarchy the same way they are stored in the file-system or in egg files.
    \item Pickled data, stored an an individual file. Class references inside the pickled files are resolved using the source in the archive.
    \item Tensor data files, stored in the ic{data/} folder with the contents of tensors.
    \item \ic{external_modules}, a special file that lists the Python modules that should be resolved using the system's builtin in module import system rather than from the package.
\end{itemize}

Figure~\ref{packaging_zip} shows what this layout looks like for the Tacotron 2 application.

We provide a \ic{PackageExporter} that helps create archives in this format, and a \ic{PackageImporter} that loads them in a hermetic way.
\subsubsection{Exporting models}
Figure~\ref{example_export} shows example code for export a model using our packaging format. It uses the method \ic{save_pickle} to save a pickled version of the file along with some annotations to describe how to treat dependencies we describe below.

In normal PyTorch, it is possible to serialize a \ic{torch.nn.Module} using the \ic{torch.save} function. This saves object using Python's pickle protocol for the Python objects, and separate files for the tensor data combined into a single zip file. Python's pickle format saves objects, but does not save the \emph{code} used to define the behavior of objects. For example if a user saves an object of class \ic{my_package.MyClass} which has an attribute \ic{my_object.my_int = 1}, then Pickle will write out bytecode in the pickling language that says ``create a \ic{my_package.MyClass} object, setting its attributes to \mintinline{Python}|{'my_int': 1}|''.  When loading the pickled object the system will use Pythons \ic{import} infrastructure to load \ic{my_package} to get the code for the object.

Libraries such as \ic{cloudpickle}~\cite{cloudpickle} extend the pickle format to include the definition of the class as well by serializing the Python code objects as well. However, this is not ideal for model packaging. It is specific to a particular version of Python, because it stores bytecodes which are not stable. More importantly, it does not produce a human-readable form of the code being exported, making attempts to edit or debug that code difficult.

Our packaging format takes an approach based on extracting entire source files instead. In \ic{save_pickle}, we scan the produced pickle bytecodes for references to modules. For each module we resolve it to its source file and include that source file in the archive. This approach produces human-readable source code. By keeping the structure of the code the same as when it was written, users can easily debug packaged models. Users can also fix bugs in the exported source or perform transfer learning with the model by unzipping the archive, editing the code. 

Some modules are implemented using C extensions rather than Python code. Others, such as \ic{torch}, are very large and would typically be included in the inference engine for use by all modules. These modules can be marked \emph{extern} in the exporter API to tell the package not to try to include their source, and inform the importer it is allowed to import them from the system library. Python standard library modules and \ic{torch} are treated as extern by default.

Python source files are not self-contained and almost all files include import statements that reference other modules. Our packaging format takes a semi-automated approach to discover these dependencies. Each source file is scanned for \ic{import} statements, we resolve these import statements to actual modules, and then export the code for that module as well, recursively scanning for modules in that code as well.

Because Python is a dynamically typed and very flexible language, it is not possible to guarantee the accuracy of dependency scanning. If code imports modules using the \ic{importlib} module, it can import arbitrary modules that cannot be detected. Code can also be loaded independently of the module system entirely. We provide a way to explicitly include modules that are missed by scanning. In practice, this is rare and none of our example models required this.

A much more common issue with dependency scanning is the inclusion of false dependencies. The object being serialized may only contain one class from a file, but since we work at the level of entire files, the exporter may include much more code than is required for the pickled object. Poorly organized code, such as a \ic{utils.py} file that aggregates a bunch of unrelated functionality can cause large amounts of code to be included that the model does not actually need. Furthermore even within a class such as a particular type of \ic{torch.nn.Module}, there might be code related to initializing the object or performing training that is not actually needed for inference. We have seen examples where this code then relies on modules unrelated to inference such as data loading or check-pointing code.

It is always possible to mitigate the discovery of false dependencies by refactoring files like \ic{utils.py} into smaller independent components and to move functionality for data loading and check-pointing out of the \ic{torch.nn.Module} classes in the Model. However, this refactoring might take significant effort, or these classes might exist in a library that the person packaging cannot easily edit. For these reasons, we provide the concept of a \emph{mocked} module in the packaging API. Before exporting a package, certain modules can be marked as mocked, which will replace their implementation in the package with a stub. Importing the module will succeed, but any attributes accessed (e.g. \ic{my_module.my_method}) will return dummy \ic{MockObjects} that will throw an exceptions when used. This allows statements like \ic{from my_module import my_method} to succeed even through the module code is not present.  Mocks can be used to manually eliminate the false dependencies for a model. Our packager provides a verbose interface to help visualize and debug the export process to make it clear what modules should be mocked. Our evaluation section describes our experience packaging models using mocking in our benchmark suite.

\begin{figure*}
\begin{tabular}{ l l l }
& \emph{Description} & \emph{Packaging Notes}\\
\emph{\href{https://github.com/avinashpaliwal/Super-SloMo}{Super SloMo}}~\cite{Super_SloMo} & Video frame interpolation &  \\
\emph{\href{https://github.com/jadore801120/attention-is-all-you-need-pytorch}{Attention is All You Need}}~\cite{attention_is_all_you_need_pytorch} & Constituency parsing &  \\
\emph{\href{https://github.com/yunjey/stargan}{Star GAN}}~\cite{pytorch_stargan} & Image to image translation &  \\
\emph{\href{https://github.com/robieta/demucs}{Demucs}}~\cite{demucs} & Music source separation & 1 mock \\
\emph{\href{https://github.com/ultralytics/yolov3}{Yolo v3}}~\cite{yolov3} & Real-time object detection & 1 mock \\
\emph{\href{https://github.com/codertimo/BERT-pytorch}{BERT}}~\cite{BERT_pytorch} & Cross-lingual natural language inference &  \\
\emph{\href{https://github.com/hzwer/LearningToPaint}{Learning To Paint}}~\cite{LearningToPaint} & Deep reinforcement learning for painting &  \\
\emph{\href{http://github.com/pytorch/vision}{MobileNet v3}}~\cite{pytorch_mobilenet_v3} & Object detection and segmentation on mobile &  \\
\emph{\href{https://github.com/harvardnlp/pytorch-struct}{Struct}}~\cite{pytorch_struct} & Structure prediction &  \\
\emph{\href{https://github.com/facebookresearch/dlrm}{DLRM}}~\cite{dlrm} & Click-through rate prediction & 3 mocks \\
\emph{\href{https://github.com/senguptaumd/Background-Matting}{Background Matting}}~\cite{Background_Matting} & Image matting without a green screen &  \\
\emph{\href{https://github.com/facebookresearch/moco}{MoCo}}~\cite{moco} & Self-supervised image classification &  \\
\emph{\href{https://github.com/facebookresearch/maskrcnn-benchmark}{Mask R-CNN}}~\cite{maskrcnn_benchmark} & Image segmentation & 4 mocks, 4 line \\ & & patch, custom kernels \\
\emph{\href{https://github.com/junyanz/pytorch-CycleGAN-and-pix2pix}{CycleGAN}}~\cite{pytorch_CycleGAN_and_pix2pix} & Image-to-image translation &  \\
\emph{\href{https://github.com/NVIDIA/tacotron2}{Tacotron 2}}~\cite{tacotron2} & Speech synthesis & 2 mocks \\
\end{tabular}
\caption{Example models used in our benchmarks. Each model was packaged from popular Github repositories of PyTorch models and is based on a published model designs. Packaging nodes describes any modifications we made to get the models exported in our package format.}
\label{models}
\end{figure*}

\subsubsection{Importing models}
The API for importing modules is shown in Listing~\ref{importing_api} and mirrors the export API. Models for similar domains will often have types of the same name. For instance, two models that contain a ResNet trunk may both contain classes called \ic{models.Resnet} with different implementations. \emph{Hermetic} model loading ensures that both of these models can be loaded into the same Python interpreter without their classes interfering with each other.

We achieve this with customization to the Python unpickler used when loading pickled data, and a custom implementation of the import infrastructure that knows how to resolve modules to the contents of a package.

In Python, the table \ic{sys.modules} holds a global view of the loaded Python modules in the system. For package code, we instead have a package-specific view of modules stored in \ic{the_importer.modules} that manages the code for package objects.

When pickled data is loaded from the archive, we use an unpickler with a modified way of resolving global references that uses \ic{the_importer.import_module} rather than the global \ic{importlib} to resolve references.  If a module is in the \ic{extern_modules} file, then the package importer uses the system importer to resolve the module. Otherwise it is loaded from the package.

When a code is loaded from the package, we install a custom \ic{__import__} method in the builtins table of the module. This change causes all \ic{import} statements to use the package-specific import process rather than the global one internal to the package.

Once an object is loaded, users can interact with it as if it were imported normally. The only difference is that the qualified name of the class's module will not match what is in \ic{sys.modules}. In most circumstances this does not affect functionality. The one place where it does is when trying to re-export objects imported from a package. The Pickler normally checks that the module of the class matches the global one to ensure an object will unpickle correctly. To overcome this limitation, our \emph{exporter} optionally takes an ordered list of importers (including the system importer) that it searches to find the implementation of any classes it pickles.

The \ic{torch::deploy::Package} object in our C++ API serves as a wrapper around this Python import infrastructure.

\section{Evaluation}
\begin{figure*}[p!]
\hspace{.5in}
\includegraphics[width=5.8in]{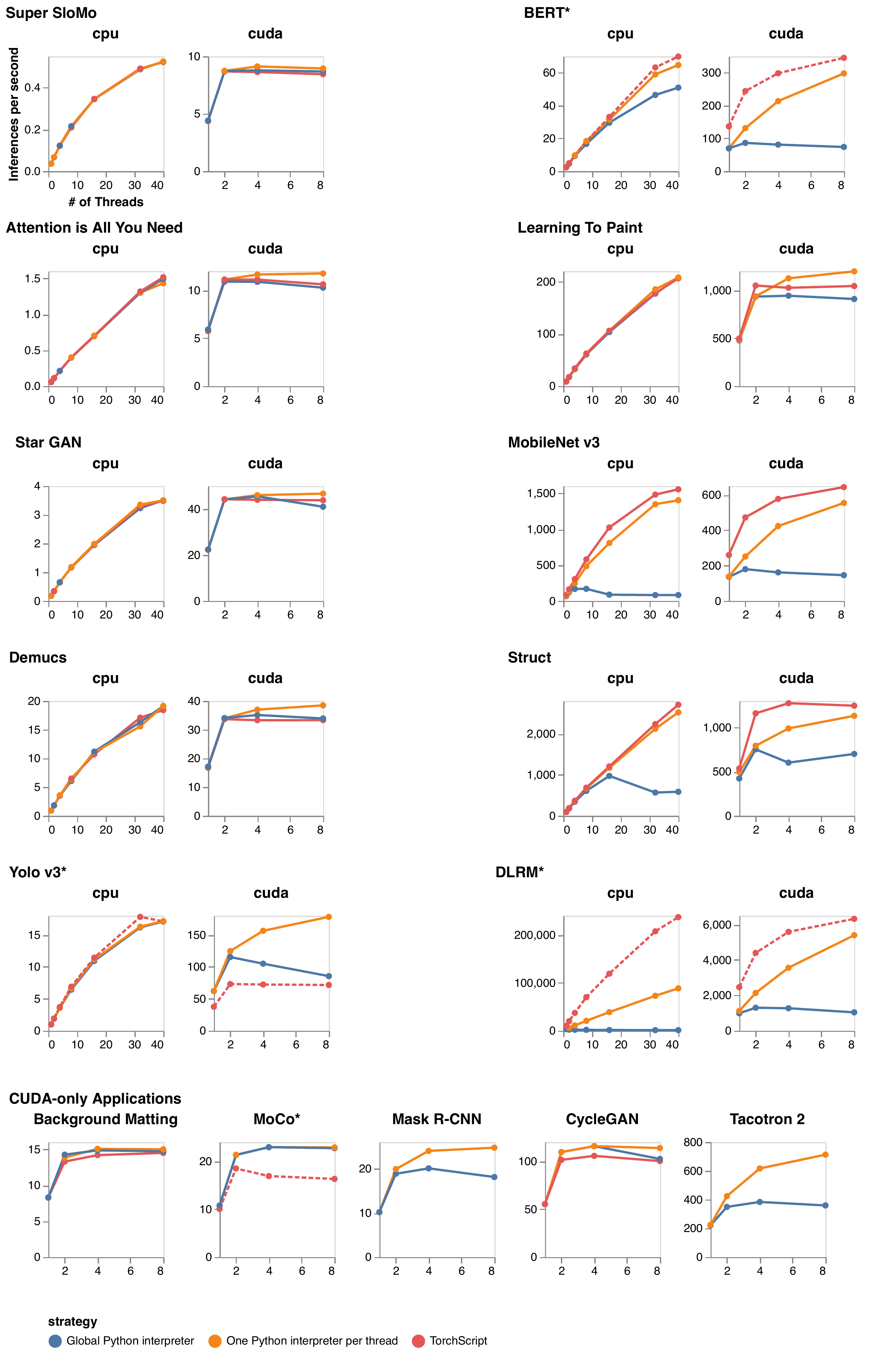}
\caption{Performance comparison of our approach (one Python interpreter per thread) against TorchScript, and a global Python interpreter, which is the way Python is typically used.}
\label{results}
\end{figure*}

By eliminating the need to extract models from Python, our system should make it faster to deploy models. However, this faster deployment is not practical if the performance of the resulting models is poor compared to the traditional approach of exporting the model. To evaluate the experience of using Python-based model inference, we assembled a suite of models, packaged them with \ic{torch.package} and compared their performance to the same models converted to TorchScript when possible. Experiments were run on a 40 core machine (2 sockets, each with 20 core Intel Xeon E5-2698V4 processors) and 2 GP100 NVIDIA GPUs using Python 3.8 and a nightly build of PyTorch from August 2020.

\subsection{The models}
Figure~\ref{models} provides a description of the models we use for our evaluation. These models are part of PyTorch's model-level benchmark suite~\footnote{https://github.com/pytorch/benchmark} and were adapted from popular Github repositories (by star count) containing PyTorch code. Rather than include popular ``trunk'' networks like ResNet in our evaluation we chose to focus on end-to-end models that include these trunks as components because we believe it provides a more accurate picture of how models will be used in practice. The benchmarks contain a number of image processing examples in additional to several models from other domains including language (BERT), audio (Demucs), speech synthesis (Tacotron 2), structure prediction (Struct), and video (Super SloMo). When preparing the models for use in our evaluation, we modified their organization so that could build within the same Python environment and provided a consistent API for loading the model, but we avoided making changes to how the code was organized that would change their performance or the ease with which they were packaged. 

\subsection{Packing pre-existing model code.}
We evaluated the usability of the \ic{torch.package} system by writing export code for each model. Export code appears similar to the code showed for the Tacotron 2 model in Figure~\ref{example_export}. By default we marked \ic{torch} as extern, and mocked out \ic{numpy} and \ic{scipy} since these were commonly included for debugging but unused in inference.  Despite being real repositories of model code used in research by a variety of authors, the models were easy to package. Seven of the fifteen models required no additional annotations.  The remaining models required a small number of mocked out libraries as  described in Figure~\ref{models}.  The most complicated model to package was MaskRCNN, which required 4 mocks and a stub replacement for the \ic{sys} module because code in the library was examining version information provided by the module. It also required additional kernels for regions of interest that are typically part of \ic{torchvision} but not the main PyTorch codebase. Section~\ref{conclusion} discusses how we can make the process of including additional per-model kernels easier. Once exported with correct mock annotations, each model was tested for correctness by comparing its results to the non-packaged version of the model for consistency.

As part of the effort of assembling the benchmarks, we had PyTorch developers add annotations to some of the model to make them amenable to TorchScript. This process was by far the most time consuming part of preparing the benchmark, with individual developers often spending on the order of several hours to make models able to be exported from Python using TorchScript. Several of our more complicated models (BERT, Yolo v3, DLRM, Mask R-CNN, and Tacotron 2) do not have TorchScript versions because of the complexity of porting them. Part of the difficulty is that TorchScript enforces static typing of the program to be able to perform more aggressive optimizations, but these changes require refactoring models to fit the type system. The process of packaging the models using \ic{torch.package} was qualitatively simpler than preparing models for TorchScript since it mostly involved breaking false dependencies using mocking.

\subsection{Performance of deployed Python models}
To measure the performance of Python-deployed models, we created a benchmark using our \ic{torch::deploy} API that simulates the model inference process. A model is loaded from the package and a number of requester threads are created that make requests to the model using example inputs. We tested both CPU and GPU inference. Since there are only 2 GPUs, when the number of threads exceeded 2, we multiplexed the use of the GPUs across threads the GPUs. For CPU inference, we instructed PyTorch to disable intra-op parallelism with \ic{OMP_NUM_THREADS=1} as is recommended for inference settings when multiple requests will provide parallelism. In this setup, ideal scaling would be linear for CPU inference up to 40 threads. On the GPU, ideal scaling would be linear up to 2 threads, and then level off as additional threads multiplex the GPUs.

Figure~\ref{results} presents the results of the benchmark in three configurations. \emph{One Python interpreter per thread} shows the performance of our approach, using the ability to load multiple python interpreters described in Section~\ref{multi_python}. To simulate what would happen without customizing Python, the \emph{Global Python Interpreter} approach limits the total number of interpreters to 1, similar to how Python is normally used. Finally as a comparison to how the models would perform when extracted from Python entirely we measured the performance of the \emph{TorchScript}-converted models where possible. BERT, DLRM, and MoCo could not be converted to TorchScript, so we used \ic{torch.jit.trace} to get a trace of computation for a particular example input. Each of these models includes some control-flow that is not represented in the trace, so these numbers server as a \emph{upper bound} on the throughput as TorchScript and may perform slower in TorchScript when fully ported. Some models did not include CPU-only models, so we only include CUDA throughput.

In each group, models are ordered by throughput, with slower (bigger) models first. Performance results for these bigger models (e.g., Super SloMo, Attention is All You Need, Star GAN, Demucs, Yolo v3) is almost the same across all three configurations. This reinforces our intuition that large Tensor operations will make Python runtime only a small component of overall time. For these models, extracting from Python via TorchScript provides little benefit as does using multiple Python interpreters. However, even for these examples having a Python packaging system to make hermetic versions of the model, and a consistent API for running the models from C++ is beneficial. 

Medium-sized models show different performance characteristics. For instance, MobileNet v3 on the CPU shows good scaling for both TorchScript and multiple Python interpreters each but the single Python interpreter barely scales showing how the global interpreter lock prevents decent inference performance. On the GPU, TorchScript performs up to 1.9x faster than multiple Python interpreters, but only 1.14x better than multiple Python interpreters when using 8 requester threads. This result indicates that there is Python overhead in running this benchmark. However, by using additional CPU cores to parallelize this overhead, it is possible to reduce the overall throughput loss using GPUs while still keeping the model in Python. This approach is wasteful of CPU cores, but might be a decent tradeoff when prototyping the deployment of a model if it prevents significant engineering effort to port the model to TorchScript.

Finally, the smallest models such as DLRM show clear Python overheads with TorchScript performing more than 2x faster. Examining a profile of the DLRM example, we see that half the time is clearly spent in the Python interpreter. In these cases it would make sense to use TorchScript for deployment if possible. Nevertheless, the multiple Python approach, while slower, still scales with the number of threads, and hence offers a scalable option for deployment before putting effort into a faster TorchScript version.

\section{Conclusion}
\label{conclusion}
Our evaluation showed that performing inference in Python using multiple interpreters is a practical way of deploying many models. For models that spend significant time in the Python interpreter, the use of multiple interpreters enables scaling when the GIL would otherwise create contention. Furthermore, for GPU inference, the ability to scale the number of Python interpreters allows the Python overhead to be amortized across multiple request threads.

Our approach to Python inference still has some limitations that can be overcome with future work. Loading of third-party C++ extensions including Python bindings, such as those in Mask RCNN is difficult because CPython bindings directly refer to global symbols in the dynamic linker table. In our approach these symbols are hidden from other extensions. We work around this by recompiling our shared interpreter library with the additional extensions included, but this is more complicated than simply including the extension library with the model package. Furthermore, our approach requires copying and loading the shared interpreter library for each interpreter, duplicating code in memory. This library is 34MB large in release mode, which is acceptable for server applications where memory is plentiful, but it grows to 274MB when debug symbols are enabled.

Both the extensions and code size issues can be resolved by writing custom dynamic library loading code rather than relying on OS primitives like \ic{dlopen}. A custom loader could map the code sections of the file into different places in virtual memory while ensuring that only one real copy of the code exists. Furthermore, since the loader is responsible for resolving external symbols, custom code could resolve Python C API references to their local copies when loading extension modules. We did not pursue this approach in our experiments because it requires significantly more complicated code to parse ELF shared library files.

Finally, even for models whose performance is less using Python, Python inference gives model authors flexibility to quickly prototype and deploy models and then focus on the performance of the models when necessary rather than having to invest in upfront effort to extract the model. Because Python does not have to be entirely eliminated, it also offers a more piecemeal approach to performance. For instance using Python-based libraries like Halide~\cite{halide} or TVM~\cite{tvm} to accelerate the model while still packaging the model as a Python program. Using Python as the packaging format opens up the possibility of employing bespoke compilers and optimizers withou the burden of creating an entire packaging and deployment environment for each technology.

\bibliography{main}
\bibliographystyle{mlsys2020}

\appendix


\end{document}